\documentclass[runningheads]{llncs}
\usepackage{graphicx}
\usepackage{textcomp}
\usepackage{xcolor}
\usepackage{bm}
\usepackage{multirow}
\usepackage{subfigure}
\usepackage{algorithmic}
\usepackage[ruled,lined,linesnumbered]{algorithm2e}
\usepackage{amsfonts}
\usepackage{amsmath}
\usepackage{booktabs}
\usepackage[colorlinks = true, linkcolor = blue, urlcolor  = blue, citecolor = blue]{hyperref}
\begin{document}
\title{SLRL: Structured Latent Representation Learning for Multi-view Clustering}
%
%
\author{Zhangci Xiong\inst{1} \and
Meng Cao\inst{2}}
\authorrunning{Z. Xiong et al.}
%
\institute{Tongji University, Shanghai, China \email{moreonenight@tongji.edu.cn} \and
Tianjin University, Tianjin, China
\email{caomeng997@tju.edu.cn}}
\maketitle              
\begin{abstract}
In recent years, Multi-View Clustering (MVC) has attracted increasing attention for its potential to reduce the annotation burden associated with large datasets. The aim of MVC is to exploit the inherent consistency and complementarity among different views, thereby integrating information from multiple perspectives to improve clustering outcomes.

Despite extensive research in MVC, most existing methods focus predominantly on harnessing complementary information across views to enhance clustering effectiveness, often neglecting the structural information among samples, which is crucial for exploring sample correlations. To address this gap, we introduce a novel framework, termed Structured Latent Representation Learning based Multi-View Clustering method (SLRL). SLRL leverages both the complementary and structural information. Initially, it learns a common latent representation for all views. Subsequently, to exploit the structural information among samples, a k-nearest neighbor graph is constructed from this common latent representation. This graph facilitates enhanced sample interaction through graph learning techniques, leading to a structured latent representation optimized for clustering. Extensive experiments demonstrate that SLRL not only competes well with existing methods but also sets new benchmarks in various multi-view datasets.

\keywords{Multi-view Clustering \and Latent
Representation \and k-nearest Neighbor Graph \and Graph Representation Learning.}
\end{abstract}
\section{Introduction}
Multi-View Clustering (MVC) has attracted significant attention in recent years, primarily because multi-view or multi-modal data offer shared semantics that enhance learning effectiveness \cite{cai2013multi,chao2021survey,chen2020multi,zhang2019multi}. This technique is especially beneficial in scenarios lacking explicit labels.

There have been notable advancements in MVC methodologies, including the development of subspace-based, non-negative matrix factorization-based, and graph-based clustering algorithms. While these methods have yielded satisfactory results, challenges remain. Specifically, most approaches concentrate on harnessing the consistency and complementarity across views to amalgamate information from diverse sources. However, they often neglect the structural information among samples, which is crucial for uncovering latent relationships essential for effective clustering. Although some graph-based MVC methods have been introduced, they typically depend on constructing graphs from original data matrices or shallow features and subsequently fusing these graphs through various strategies. This method is not only computationally demanding but also heavily reliant on the quality of the initial graph construction, thus potentially limiting the effectiveness of MVC.

\begin{figure}[ht!]
	\centering
	\includegraphics[width=4.5in,height = 1.5in]{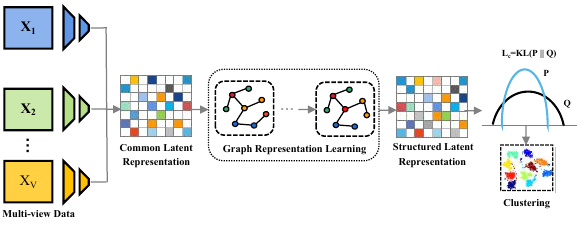}
	\caption{SLRL model framework.}
	\label{SLRL_framework}
\end{figure}

In response to these issues, this paper proposes a \textbf{Structured Latent Representation Learning} (\textbf{SLRL}) based multi-view clustering method. The model structure of SLRL is illustrated in Figure \ref{SLRL_framework}. SLRL initially mines the complementary information of different views flexibly by learning a common latent representation. Subsequently, to leverage the structural information among samples, a $k$-nearest neighbor graph is constructed based on this latent representation. This graph is further enhanced through a graph attention network, thereby learning a structured latent representation. Moreover, a clustering loss further constrains the latent representation learning, endowing the learned latent representation with a favorable cluster structure. Compared with existing multi-view clustering methods, SLRL has the following advantages:

(1) By considering both the complementary information between views and the structural relationships among samples, the proposed method not only flexibly integrates information from multiple views but also fully utilizes the structural information among samples.

(2) The learned latent representations are not only well-structured but also cohesive, making them particularly suitable for clustering tasks.

(3) Experiments conducted on six multi-view datasets demonstrate that the clustering performance of SLRL surpasses that of several existing methods.
\section{Related Work}

\subsection{Multi-view Clustering}

Multi-view clustering aims to achieve enhanced clustering performance by integrating information from different views. One straightforward approach involves concatenating features from all views into a new feature vector, followed by the application of traditional single-view clustering methods such as K-means or spectral clustering on this vector. However, this strategy often falls short of expectations as it overlooks the distinct characteristics of each view and the complex relationships between them, and it lacks interpretability. To better extract information from multi-view data, multi-view clustering algorithms generally adhere to two principles: the principle of consistency and the principle of complementarity. In recent years, numerous multi-view clustering methods have been proposed by effectively considering the consistency and complementarity of information from multiple views\cite{kumar2011co,zhan2018multiview,zhan2017graph,Zhang-et-al:2018,zhang2017flexible,zhang2017latent,xia2014robust}. The existing multi-view clustering methods can be categorized into four types: co-training based multi-view clustering, multi-kernel learning based multi-view clustering, multi-view subspace clustering, and graph-based multi-view clustering. Our work pertains to graph-based multi-view clustering.

\subsection{Graph-based Multi-view Clustering}
Graph-based multi-view clustering firstly fuses all views into a single graph, and then applies graph cutting algorithms or spectral clustering on the fused graph to obtain the final clustering results.

In recent years, numerous graph learning-based multi-view clustering methods have been proposed. For instance, Saha et al.\cite{saha2013graph} introduced a graph-based multi-view clustering method that initially employs three different similarity measures to construct nearest neighbor graphs projected onto individual feature spaces. Subsequently, these graphs are clustered into feature subspaces, followed by hierarchical agglomerative clustering on each view to achieve the clustering outcome. Nie et al.\cite{Nie2017} proposed a method that combines multi-view clustering and semi-supervised classification. This approach involves adaptively selecting neighbors of samples to construct graphs, and then conducting multi-view learning based on these graphs, thereby realizing more accurate clustering and classification. Tao et al.\cite{tao2017multi} introduced a multi-view clustering method based on adaptive learning of graphs, which not only considers the importance of each view at the view level but also introduces sample-pair-specific weights within views, thereby adaptively learning a common similarity matrix suitable for clustering. To enhance multi-view clustering performance, Zhan et al.\cite{zhan2017graph} proposed a graph learning based clustering method, which learns a unified graph with rank constraints on its Laplacian matrix from different perspectives. Then, clustering indicators are directly obtained from the global graph, eliminating the need for any graph cutting or K-means clustering. Tang et al.\cite{tang2020cgd} employed a Cross-view Graph Diffusion (CGD) approach, which takes predefined graph matrices from different views as input and learns a unified graph for multi-view data clustering through an iterative cross-diffusion process. Li et al.\cite{Li2022consensus} presented a consensus graph learning-based multi-view clustering method. This method integrates graph structural information from different views to construct a consensus graph, aiming to enhance clustering performance.

\section{The Proposed Method}

As shown in Figure \ref{SLRL_framework}, the SLRL method proposed in this paper is primarily composed of three parts: common latent representation learning, structured latent representation learning and clustering.

\subsection{Common Latent Representation Learning}

The essence of multi-view clustering is the effective use of consistency and complementarity between views for a comprehensive information set. Therefore, it requires integrating information from multiple views via specific strategies. Addressing view complementarity, Zhang et al.\cite{zhang2017latent} introduced a latent multi-view subspace clustering approach. This approach generates a common latent representation for all views and reconstructs each view from it. Thanks to the multi-view data's complementarity, this representation is more complete, accurate, and robust than that from a single view.

Specifically, for the $v$ view $\bm{X}_{n}^{(v)}$, define the latent representation of the $n^{th}$ sample as $h_n$; let {\small $f_{v}\left(\bm{h}_{n} ;
\bm{\Theta}_{r}^{(v)}\right)$} be the reconstruction network for the $v^{th}$ view, parameterized by {\small $\Theta_{r}^{(v)}$}. The optimization objective for learning the common latent representation can then be defined as follows:

\begin{equation}\label{Lr}
L_{r}=\sum_{v=1}^{V}
\left\|f_{v}\left(\bm{h}_{n} ;
\bm{\Theta}_{r}^{(v)}\right)-\bm{x}_{n}^{(v)}\right\|^{2}
,
\end{equation}
where $L_{r}$ represents the reconstruction loss for the common latent representation, which aims to learn the mapping between the original data space and the common latent space. In this paper, we utilize a multi-layer perceptron as the encoder to achieve the mapping between the common latent representation and the original features $x_n$ of data. By minimizing the loss corresponding to the formula \eqref{Lr}, a common latent representation $\bm{H}=$ $\left\{\bm{h}_{1}, \bm{h}_{2}, \ldots, \bm{h}_{N}\right\}$ can be learned.

\subsection{Structured Latent Representation Learning} 

While the common latent representation can integrate complementary multi-view information, it often lacks a well-defined cluster structure, making it suboptimal for direct clustering use. Additionally, this perspective neglects the structural information among samples, which is vital for clustering. Noting that samples from the same class show similar features, we select the $k$-nearest neighbors for each sample. Thus, we propose constructing an undirected $k$-nearest neighbor graph on $\bm{H}$ to capture structural information in the latent space, defining the neighbor graph as $\bm{A}\in\bm{R}^{N*N}$. The distance between any two samples $i$ and $j$ with representations $\bm{h}_{i}$ and $\bm{h}_{j}$ is calculated as follows.

For continuous data such as images, we employ a Gaussian kernel to measure the similarity between sample $i$ and sample $j$. The neighbor relationship between samples $i$ and $j$ can be calculated using the following formula:

\begin{equation}
	\centering
	\mathrm{A}_{i j}=\left\{\begin{array}{l}
	\exp \left(-\frac{\left\|\bm{h}_{i}-\bm{h}_{j}\right\|^{2}}{2 \sigma^{2}}\right), \bm{h}_{i} \in N_{k}\left(\bm{h}_{j}\right) \text { or }~ \bm{h}_{j} \in N_{k}\left(\bm{h}_{i}\right), \\
	0, \text {other},
	\end{array}\right.
    \label{distance1}
\end{equation}
For discrete data, such as text, we utilize the dot product similarity for calculating the similarity between entities. The neighbor relationship can then be computed as follows:

\begin{equation}
	\centering
	\mathrm{A}_{i j}=\left\{\begin{array}{l}
	\bm{h}_{j}^{T}\bm{h}_{i},\bm{h}_{i}\in N_{k}\left(\bm{h}_{j}\right) \text {or}~ \bm{h}_{j} \in N_{k}\left(\bm{h}_{i}\right), \\
	0, \text {other},
	\end{array}\right.
	 \label{distance2}	
\end{equation}
where $\bm{N}_{k}\left(\bm{h}_{i}\right)$ denotes the $k$-nearest neighbors of the $i^{th}$ sample. Thus, we can obtain the neighbor graph $\bm{A}$ from non-graph data, which effectively represents the neighbor relationships among samples.

Given the effectiveness of nearest neighbor graphs in assessing distances between diverse samples, we leverage \textbf{Graph Attention Networks} (\textbf{GAT}) to explore the information among samples within the common latent representation, thereby obtaining a structured latent representation.

The GAT takes $\bm{H}=$ $\left\{\bm{h}_{1}, \bm{h}_{2}, \ldots, \bm{h}_{N}\right\}$ as input, and the output of the GAT node features is denoted as $\tilde{\bm{h}_{i}} \in \mathbb{R}^{F^{\prime}}$, where $F$ represents the dimensionality of the input node features, and $F^{\prime}$ denotes the dimensionality of the output node features. The objective of GAT is to achieve sufficient expressive capability by transforming the input features into higher-dimensional features. Initially, it applies a learnable linear transformation to each node, with the weight matrix represented as $\bm{W} \in \mathbb{R}^{F^{\prime}\times F} $. Subsequently, it employs a single-layer feed-forward neural network, denoted as $\bm{a} \in \mathbb{R}^{2 F^{\prime}}$, to calculate the attention coefficients between nodes. The process of computation is illustrated in Equation (\ref{attention}).

\begin{equation}
{\alpha}_{i j}=\frac{\exp \left(\operatorname{LeakyReLU}\left(\bm{a}^{T}\left[\mathbf{W} \bm{h}_{i} \| \bm{W} \bm{h}_{j}\right]\right)\right)}{\sum_{j \in \mathcal{N}_{i}} \exp \left(\operatorname{LeakyReLU}\left(\bm{a}^{T}\left[\mathbf{W} \bm{h}_{i} \| \mathbf{W} \bm{h}_{j}\right]\right)\right)}
,
\label{attention}
\end{equation}
where $T$ denotes the matrix transpose, and $\|$ represents the concatenation operation. Node $j$ is a neighbor of node $i$, and $\alpha_{i j}$ signifies the importance of the features of node $j$ relative to node $i$ among all the neighbors of node $i$. The set of all neighbors of node $i$ is represented by $\mathcal{N}_{i}$.

Utilizing the normalized attention coefficients $\alpha_{i j}$, GAT combines the features of neighboring nodes to update the current node's features, then applying the nonlinear sigmoid function $\sigma$. The computation of the node features is as follows:

\begin{equation}
\tilde{\bm{h}_{i}}=\sigma\left(\sum_{j \in \mathcal{N}_{i}} \alpha_{ij} \bm{W} \bm{h}_{j}\right)
.
\end{equation}
Finally, GAT stabilizes the learning process by employing multi-head attention, and concatenates or averages the output features of the node's neighbors to form its final output feature, as illustrated in Equation (\ref{multi-head}):

\begin{equation}
\tilde{\bm{h}_{i}}=\|_{k=1}^{K} \sigma\left(\sum_{j \in \mathcal{N}_{i}} \alpha_{i j}^{k} \bm{W} \tilde{\bm{h}_{i}}\right)
,
\label{multi-head}
\end{equation}
where $K$ denotes the number of attention heads, and $\alpha_{i j}^{k}$ represents the relative attention weight of the features of node $j$ with respect to node $i$ for the $k^{th}$ attention head.

By applying attention mechanism learning to each node, we get a structured latent representation ${\tilde{\bm{H}}}=\left\{\tilde{\bm{h}_{1}},\tilde{\bm{h}_{2}},  \ldots, \tilde{\bm{h}_{N}}\right\}$. For each sample in $\tilde{\bm{H}}$, similar samples feature closer characteristics during the learning process, resulting in an enhanced structural integrity.

\subsection{Joint Training and Clustering}

Through GAT, we obtain a structured latent representation $\tilde{\bm{H}}$ that preserves data's inherent local structures but isn't directly suitable for clustering. Typically, existing works apply traditional clustering methods like spectral clustering or K-means to these features, with clustering effectiveness dependent on the feature quality. To address this, Xie et al. \cite{xie2016unsupervised} introduced an end-to-end clustering method that simultaneously optimizes latent representation learning and clustering, enhancing the suitability of the structured latent representation for clustering tasks.

In this paper, we use a two-step alternating unsupervised algorithm to improve clustering. Initially, we compute the soft assignment of each sample in the latent representation to cluster centroids. Next, we update the structured latent representation and refine centroids based on high-confidence assignments using an auxiliary target distribution. This cycle repeats until convergence. Specifically, for each $i^{th}$ sample and $j^{th}$ cluster, similarity between $\tilde{\bm{h}}{i}$ and $\bm{u}_{j}$ is measured using a $t$-distribution kernel, calculated as follows:

\begin{equation}
    q_{i j}=\frac{\left(1+\left\|{\tilde{{\bm{h}_i}}}-\bm{\mu}_{j}\right\|^{2}\right)^{-1}}{\sum_{j}\left(1+\left\|{\tilde{\bm{h}}_{i}}-\bm{\mu}_{j}\right\|^{2}\right)^{-1}}
    \label{calculate_q}
    .
\end{equation}
The cluster centers $u_{i}$ are initialized through K-means on $\tilde{{\bm{h}_i}}$. $Q={[q_{i j}]}$ represents the probability of assigning sample $i$ to cluster $j$, i.e. the soft assignment. Once the clustering distribution $Q$ is obtained, our goal is to optimize the latent representation through learning from high-confidence soft assignments. Specifically, we aim to bring the feature representation of each sample closer to its cluster center, thereby enhancing the cohesiveness within the cluster. Therefore, the $p_{ij}$ in the target distribution $P$ can be calculated as follows:

\begin{equation}
    p_{i j}=\frac{q_{i j}^{2} / \sum_{i} q_{i j}}{\sum_{j}\left(q_{i j}^{2} / \sum_{i} q_{i j}\right)}
    \label{calculate_p}
    .
\end{equation}
Within the target distribution $P$, each assignment in $Q$ is squared and normalized to confer a higher confidence level. Consequently, the clustering loss function can be defined as follows:

\begin{equation}
    L_c=KL(P \| Q)=\sum_{i} \sum_{j} p_{i j} \log \frac{p_{i j}}{q_{i j}}
    .
\end{equation}
By minimizing the KL divergence loss between the distributions $Q$ and $P$, the target distribution $P$ facilitates the network in learning a structured latent representation that is more suitable for clustering tasks. This implies that data representations around the cluster centers are brought closer together, thereby yielding improved clustering performance.

\begin{algorithm}[!ht]
	\caption{The Training Process of SLRL.}
	\textbf{Input}: The multi-view data $\bm{X}=\left\{{\bm{X}}^{(1)},{\bm{X}}^{(2)},...,{\bm{X}}^{(V)}\right\}$, the number of nearest neighbors $k$, the hyperparameter $\gamma$, and the maximum number of iterations $T$:
		\\
		\textbf{Initialization}: Randomly initialize
		$\left\{\bm{h}_{n}\right\}_{n=1}^{N}$, $\left\{\bm{\Theta}_{r}^{(v)}\right\}_{v=1}^{V}$ and $\left\{\bm{W}_{l}\right\}_{l=1}^{L}$.
	
	\For{i $\in 0,1, \cdots, T$}
	{
		Update the common latent representation $H$ using Equation (\ref{Lr});\\
	    Construct the $k$-nearest neighbor graph using either Equation (\ref{distance1}) or Equation (\ref{distance2});\\
	    \For{$l=1$ to $L$}
		{
			Update the latent representation $\bm{H}^{(\ell)}$ for layer $l$ using Equation (\ref{multi-head});\\
		}
        Calculate the reconstruction loss $L_{r}$ and the clustering loss $L_{c}$ separately, and compute the total loss using Equation (\ref{total-loss}); \\
        Update network parameters through backpropagation.
	}
	
	\textbf{Output: } Network parameters $\left\{\bm{\Theta}_{r}^{(v)}\right\}_{v=1}^{V}$, $\left\{\bm{W}_{l}\right\}_{l=1}^{L}$, structured latent representation ${\tilde{\bm{H}}}$.
	\label{slrl}
\end{algorithm}

\subsection{Loss Function}

For SLRL, the loss function comprises reconstruction loss and clustering loss. The reconstruction loss is utilized to constrain the error between the features reconstructed from the common latent representation $\bm{H}$ and the original features of the view, thus flexibly balancing consistency and complementarity. Subsequently, by exploring the structural characteristics among samples, a structured latent representation $\tilde{\bm{H}}$ is obtained. The clustering loss is employed to disperse the latent representation $\tilde{\bm{H}}$, while constraining the learning of both the common latent representation and the graph representation. This approach ensures that the learned latent representation possesses a better cluster structure. Therefore, the overall loss function can be defined as:

\begin{equation}
    L=L_{r}+\gamma L_{c}
    \label{total-loss},
\end{equation}
where $L_{r}$ and $L_{c}$ represent the reconstruction loss and clustering loss, respectively. $\gamma>0$ is a hyperparameter employed to balance the optimization of clustering and the learning of the common latent representation. Algorithm 1 succinctly summarizes the training process of SLRL.

\section{Experiments}
In this section, we present comprehensive experiments to assess the efficacy of our proposed SLRL algorithm. To this end, we compare its clustering performance with other established methods, analyze the affinity matrix and convergence behavior, investigate the parameter sensitivity, and visualize the data. 

\subsection{Experimental Settings}

In this paper, six commonly used multi-view datasets were selected for comparison of all methods, including \textbf{100leaves}\footnote{https://archive.ics.uci.edu/dataset/241/one+hundred+plant+
species+leaves+data+set}, \textbf{Scene-15}, \textbf{MSRCV1\footnote{http://research.microsoft.com/en-us/projects/objectclassrecognition}}, \textbf{3Sources\footnote{http://erdos.ucd.ie/datasets/3sources.html}}, \textbf{BBCSport\footnote{http://erdos.ucd.ie/datasets/segment.html}} and \textbf{HW}\footnote{https://archive.ics.uci.edu/dataset/72/multiple+features}, and the following ten methods were selected for comparison with the approach proposed: \textbf{SPC$_{BestSV}$}\cite{NgJW01}, \textbf{LRR$_{BestSV}$}\cite{LiuLYSYM13}, \textbf{DiMSC}\cite{cao2015diversity}, \textbf{AMGL}\cite{nie2016parameter}, \textbf{MLAN}\cite{NieCL17}, \textbf{LMSC}\cite{zhang2017latent}, \textbf{MVGL}\cite{zhan2017graph}, \textbf{GMC}\cite{wang2019gmc}, \textbf{CGD}\cite{tang2020cgd} and \textbf{DealMVC}\cite{DealMVC}.

For each experiment, 80\% of the dataset is used as the training set and 20\% as the test set. To ensure fairness, for all compared methods, adjustments are made according to the source code and parameters provided by the authors of the papers to achieve the best clustering results. For our model, the ReLU function is used as the nonlinear activation function in the network. The dimensionality of the latent representation $F$ for each dataset ranges from $\left\{16,32,64,128,256\right\}$, with the default setting being 64. The number of neighbors $k\in\{3,15\}$, with a default setting of $k=10$. As the model is not sensitive to changes in the parameter $\gamma$, $\gamma$ is uniformly set to 10 in the experiments of this paper. The loss function of the model is optimized using the batch gradient descent algorithm, with a learning rate of 0.01. To reduce random errors, we repeat each experiment 10 times and calculate the average values and standard deviations of four clustering performance metrics for comparison.

\subsection{Experimental Analysis and Results}

Tables \ref{clustering-all} presents the average values and standard deviations of different metrics across all methods on six datasets. For each clustering metric, we utilize a red bold font to denote the best performance and an underlined font to signify the second-best performance. Based on these experimental results, the following conclusions can be drawn:

(1) It is evident that, across six commonly used multi-view datasets, the method proposed in this paper has achieved excellent clustering performance. Compared to the methods reviewed, the proposed method achieved the best results on four clustering metrics across the datasets of 100Leaves, Scene-15, MSRCV1, 3Sources and HW. Relative to the second-best method, SLRL improved by 9.05\%, 8.84\%, 1.54\%, 6.48\% and 5.13\% on ACC, and by 2.34\%, 25.21\%, 7.57\%, 2.18\% and 0.37\% on NMI, respectively. Although MLAN and CGD performed better on the BBCSport dataset, their performance was lower on other datasets compared to the proposed method. Overall, SLRL demonstrated desirable clustering performance across different datasets, underscoring its superiority.

(2) On some datasets, the best results of single-view clustering were superior to those of some multi-view clustering algorithms, indicating that fully utilizing the information from multiple views is challenging and requires more rational exploration of the consistency and complementarity between views.

\begin{table}[htbp]\scriptsize
\caption{Comparison of clustering performance on six datasets.}\label{clustering-all}
\centering
\renewcommand\arraystretch{0.9} {
	\setlength{\tabcolsep}{2mm}{
		\begin{tabular}{cccccc}
			\toprule[1.5pt]
{Datasets} & {Method}  & {ACC (\%)}  & {NMI (\%)}  & {F\_score (\%)}  & {ARI (\%)}  \\
\midrule[1pt]
\multirow{10}{*}{100Leaves\cite{100leaves}} & $\text{SPC}_{\text {BestSV} }$ & 48.03±1.11 & 76.61±2.18 & 40.76±1.84 & 40.19±1.51 \\
& $\text{LRR}_{\text {BestSV}}$ & 47.86±2.49 & 70.62±1.37 & 38.42±2.67 & 35.08±0.91 \\
& DiMSC&85.58±3.13 &72.34±2.31 &57.97±3.25 &60.89±2.96 \\
& AMGL & 76.76±1.93 & 87.57±2.86 & 57.66±1.30 & 56.12±1.94 \\
& MLAN&84.02±0.72 &94.11±1.30 &80.21±1.07 &79.56±2.36 \\
& LMSC&74.37±2.06 &86.41±2.73 &64.31±2.84 &63.30±2.95 \\
& MVGL&76.59±1.26 &85.87±0.78 &51.32±2.74 &51.37±0.70 \\
& GMC&83.62±0.83 &90.09±2.10 &80.94±2.65 &73.79±0.68 \\
& CGD&\underline{86.34±1.21} & \underline{94.92±1.93} & \underline{83.74±0.93} & \underline{81.86±1.02} \\
& DealMVC&80.44±5.41 &77.13±6.02 &78.72±5.90 &58.64±6.14 \\
& Ours (SLRL)& \textcolor{red}{\textbf{94.15±2.19}} & \textcolor{red}{\textbf{97.14±0.86}} & \textcolor{red}{\textbf{88.70±3.37}} & \textcolor{red}{\textbf{88.53±3.43}}\\
\midrule[0.5pt]
\multirow{10}{*}{Scene-15\cite{scene15}} & $\text{SPC}_{\text {BestSV} }$&37.42±1.34 &37.26±0.91 &28.31±1.70 &19.42±1.28 \\
& $\text{LRR}_{\text {BestSV}}$&36.29±0.92 &36.13±1.47 &25.31±0.90 &18.95±0.91 \\
& DiMSC&38.22±1.13 &39.00±2.40 & \underline{31.87±1.25} &22.91±1.17 \\
& AMGL&33.92±2.33 &37.35±1.33 &25.02±0.83 &17.39±1.10 \\
& MLAN&15.78±1.08 &16.59±0.03 &15.21±1.08 &13.94±1.37 \\
& LMSC&42.68±1.40 &34.98±2.21 &27.77±1.27 &23.26±0.93 \\
& MVGL&37.48±2.06 &37.29±2.17 &27.82±1.10 &22.14±2.06 \\
& GMC&41.55±1.30 &37.09±1.81 &30.04±0.92 &22.18±1.98 \\
& CGD&\underline{43.66±1.99} & \underline{42.53±1.53} &31.39±2.80 & \underline{24.94±1.53} \\
& DealMVC& 29.87±6.56 & 25.16±4.98 & 29.05±6.62 & 22.66±4.49 \\
& Ours (SLRL)& \textcolor{red}{\textbf{47.52±1.87}} & \textcolor{red}{\textbf{53.25±1.02}} & \textcolor{red}{\textbf{42.31±0.97}} & \textcolor{red}{\textbf{31.88±1.24}} \\
\midrule[0.5pt]
\multirow{10}{*}{MSRCV1\cite{msrcv1}} & $\text{SPC}_{\text {BestSV}}$&51.45±2.41 &50.07±0.85 &48.68±1.64 &47.70±1.89  \\
& $\text{LRR}_{\text {BestSV}}$&66.75±3.06 &59.68±1.55 &60.44±3.03 &58.24±1.17  \\
& DiMSC&74.33±2.61 &66.75±2.73 &73.79±0.68 &72.95±1.13  \\
& AMGL&74.40±2.91 &72.97±1.69 &66.50±2.42 &62.62±1.84  \\
& MLAN&71.41±1.82 &74.40±0.73 &66.73±2.66 &60.88±1.38  \\
& LMSC&83.62±1.33 &70.52±1.27 &69.96±2.58 &66.13±1.35  \\
& MVGL&90.72±2.46 &82.51±0.47 & \underline{81.21±1.50} &76.64±1.22  \\
& GMC&90.55±2.29 &82.76±2.03 &78.48±1.94 &76.95±2.27  \\
& CGD& \underline{91.29±1.09} & \underline{83.63±2.51} &80.98±0.82 & \underline{77.79±0.81} \\
& DealMVC& 76.52±6.85 & 65.01±7.60 & 74.64±6.32 & 59.25±7.84 \\
& Ours (SLRL) &\textcolor{red}{\textbf{92.70±2.12}} & \textcolor{red}{\textbf{89.96±1.52}} & \textcolor{red}{\textbf{84.35±3.09}} & \textcolor{red}{\textbf{82.26±1.84}} \\
\midrule[0.5pt]
\multirow{10}{*}{BBCSport} & $\text{SPC}_{\text {BestSV}}$&35.05±2.47 &30.71±1.09 &39.84±2.61 &31.92±1.65 \\
& $\text{LRR}_{\text {BestSV}}$&89.68±1.91 &77.34±2.90 &80.18±1.39 &73.75±2.26 \\
& DiMSC&84.98±1.11 &89.08±1.51 &91.27±1.20 &91.63±1.75 \\
& AMGL&57.83±2.49 &52.14±2.50 &56.81±1.80 &52.49±0.46 \\
& MLAN&\underline{97.23±3.13} & \underline{90.68±1.69} &\textcolor{red}{\textbf{94.77±2.94}} & \underline{92.07±3.08} \\
& LMSC&92.79±0.72 &83.77±1.78 &90.21±1.04 &86.07±2.05 \\
& MVGL&80.89±1.59 &79.90±2.38 &79.55±1.11 &74.81±2.35 \\
& GMC&85.56±2.30 &81.01±0.97 &80.85±1.26 &81.15±0.37 \\
& CGD&\textcolor{red}{\textbf{97.37±1.03}} &\textcolor{red}{\textbf{91.09±2.76}} & \underline{93.97±1.75} &\textcolor{red}{\textbf{93.73±1.20}} \\
& DealMVC&66.06±11.94 &49.87±12.67 &63.34±12.55 &43.30±14.93 \\
& Ours (SLRL) &93.02±2.52 & 86.44±0.93 & 91.20±1.73 & 90.75±1.10 \\
\midrule[0.5pt]
\multirow{10}{*}{3Sources} & $\text{SPC}_{\text {BestSV}}$&64.08±2.44 &56.25±1.36 &60.26±0.87 &53.15±0.70 \\
& $\text{LRR}_{\text {BestSV}}$&63.64±1.98 &53.73±1.25 &54.37±2.41 &46.07±0.75 \\
& DiMSC&73.75±2.22 &68.84±0.82 &70.23±1.25 &60.43±1.51 \\
& AMGL&63.88±1.08 &60.89±1.91 &62.75±1.27 &54.91±1.03 \\
& MLAN&68.72±2.41 &54.38±1.21 &55.73±1.33 &35.55±1.77 \\
& LMSC&72.22±2.08 &68.40±0.91 & \underline{70.61±1.15} &57.76±0.55 \\
& MVGL&70.54±1.66 &65.69±2.31 &60.63±1.59 &43.10±1.08 \\
& GMC&69.56±1.99 &61.84±2.33 &60.78±1.04 &44.38±2.52 \\
& CGD& \underline{76.65±1.53} &\underline{70.79±0.58} &70.18±3.23 & \underline{60.64±1.20}  \\
& DealMVC& 65.82±7.36 & 59.95±8.34 & 64.24±7.25 & 56.07±7.06 \\
& Ours (SLRL) & \textcolor{red}{\textbf{81.62±1.62}} & \textcolor{red}{\textbf{72.33±2.05}} & \textcolor{red}{\textbf{77.90±1.32}} & \textcolor{red}{\textbf{69.86±1.27}} \\ 
\midrule[0.5pt]
\multirow{10}{*}{HW} & $\text{SPC}_{\text {BestSV}}$& 46.66±1.71 & 42.08±1.63 & 29.69±1.48 & 40.11±1.67  \\
& $\text{LRR}_{\text {BestSV}}$& 45.42±0.31 & 40.64±0.17 & 27.05±0.23 & 34.76±0.20  \\
& DiMSC& 86.67±2.32 & 79.31±2.18 & 76.14±1.17 & 77.58±2.49  \\
& AMGL& 85.32±0.77 & 84.87±0.45 & 72.00±1.21 & 76.31±0.96  \\
& MLAN& \underline{96.04±1.74} & \underline{91.96±2.10} & 83.01±1.81 & 84.29±0.46  \\
& LMSC& 80.33±3.60 & 79.02±2.07 & 71.88±3.38 & 74.78±3.01 \\
& MVGL& 85.16±1.36 & 79.67±1.52 & 83.32±1.00 & 75.54±2.64  \\
& GMC& 88.12±2.41 & 91.79±1.43 & 80.10±0.88 & 86.62±1.17  \\
& CGD& 91.70±2.15 & 89.95±1.30 & \underline{85.44±3.01} & \underline{90.13±1.66}  \\
& DealMVC& 82.67±4.84 & 77.29±5.93 & 80.37±5.56 & 75.82±6.27 \\
& Ours (SLRL) & \textcolor{red}{\textbf{96.40±2.06}} & \textcolor{red}{\textbf{95.11±2.21}} & \textcolor{red}{\textbf{92.75±3.77}} & \textcolor{red}{\textbf{92.02±4.16}} \\
\bottomrule[1.5pt]
\end{tabular}
}
}
\end{table}

\subsection{Visualization of Clustering Results}
\begin{figure}[htbp!]
\centering
\subfigure[\text{$\text{LRR}_{\text {BestSV} }$}]{
\begin{minipage}[t]{0.2\linewidth}
\centering
\includegraphics[width=1\linewidth]{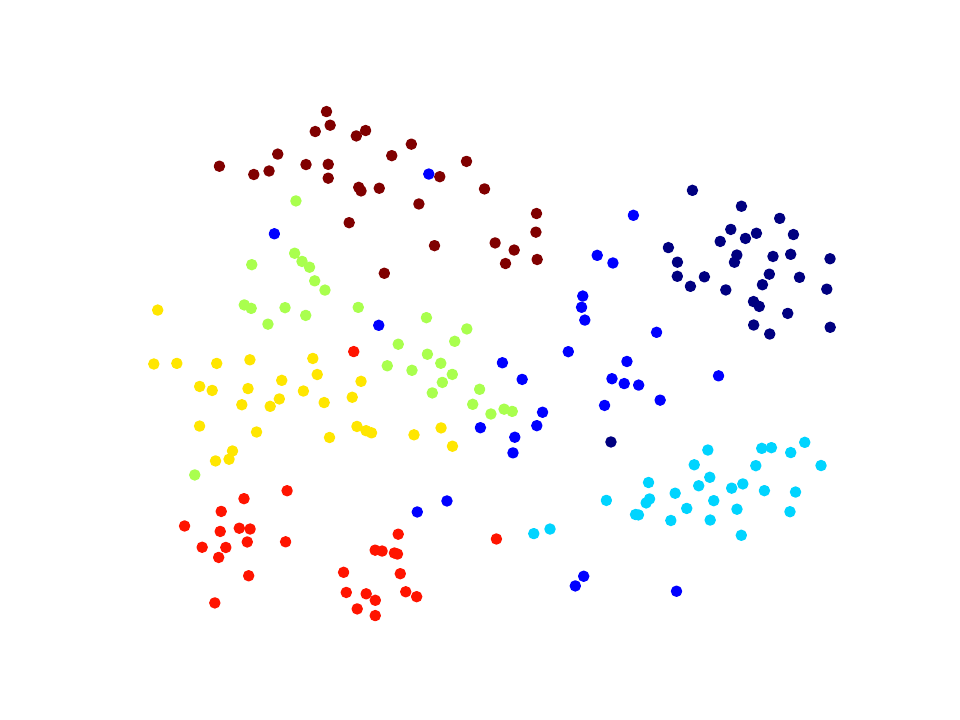}
\end{minipage}%
}%
\subfigure[\text{$\text{SPC}_{\text {BestSV} }$}]{
\begin{minipage}[t]{0.2\linewidth}
\centering
\includegraphics[width=1\linewidth]{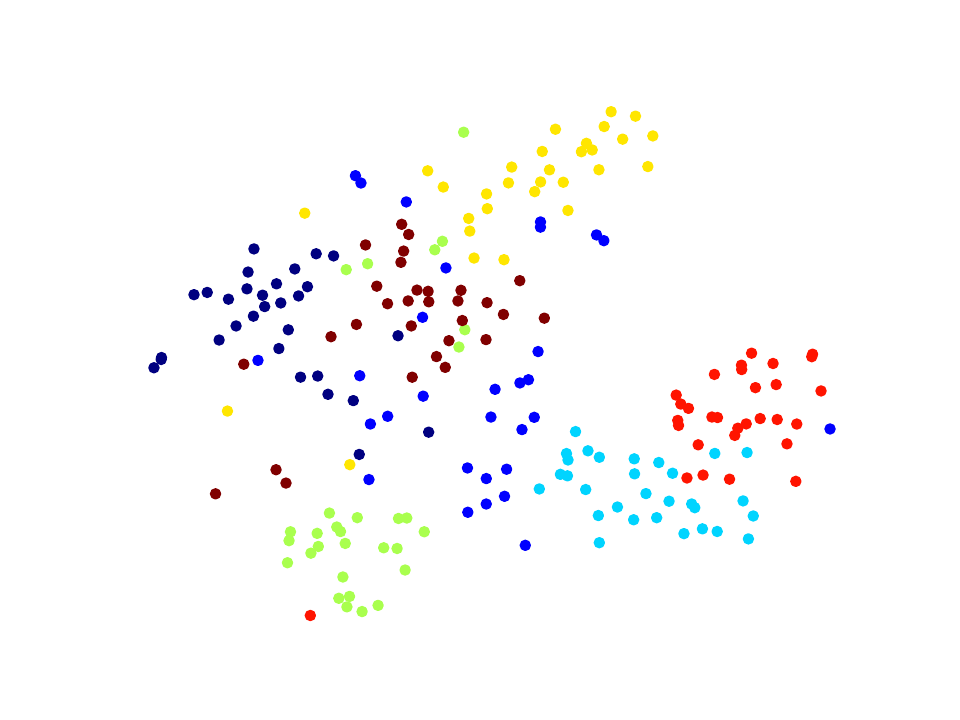}
\end{minipage}%
}%
\subfigure[MLAN]{
\begin{minipage}[t]{0.2\linewidth}
\centering
\includegraphics[width=1\linewidth]{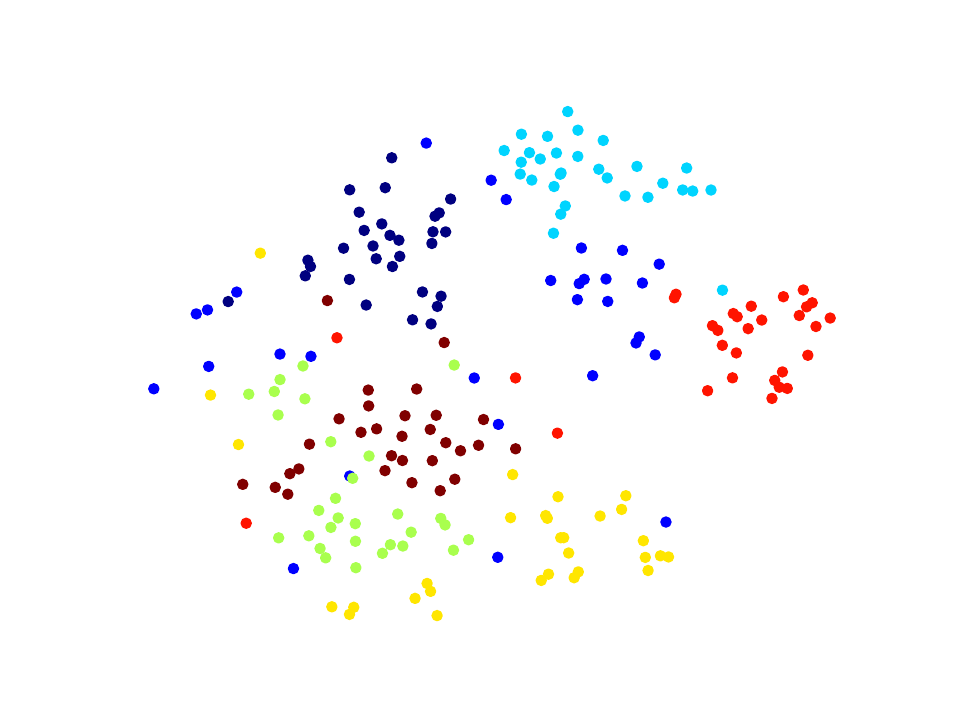}
\end{minipage}%
}%

\subfigure[LMSC]{
\begin{minipage}[t]{0.2\linewidth}
\centering
\includegraphics[width=1\linewidth]{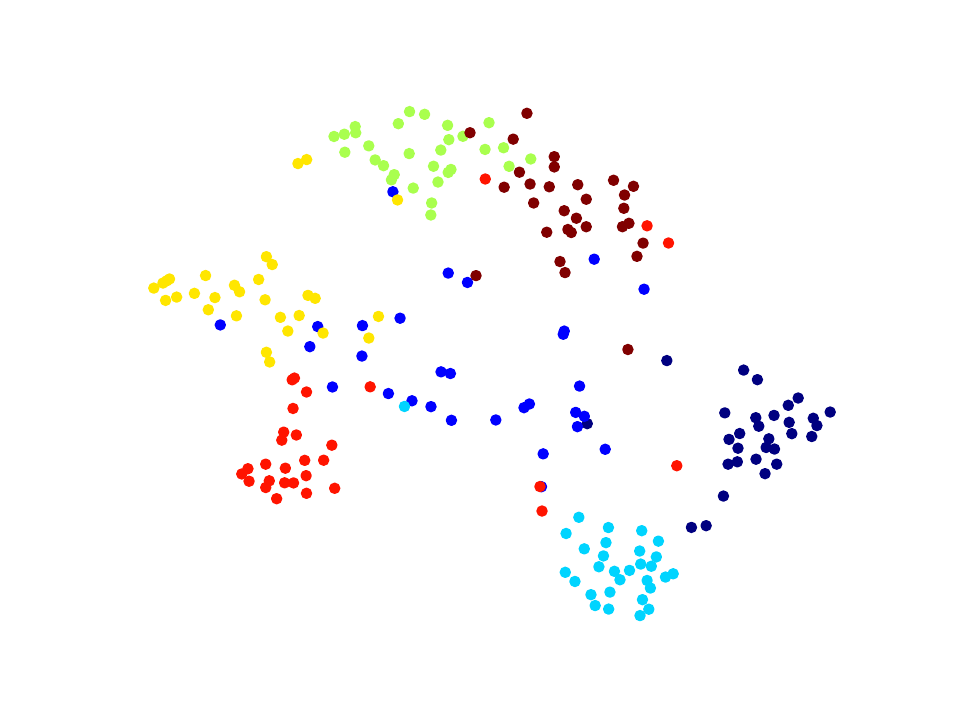}
\end{minipage}%
}%
\subfigure[SLRL-{$\bm{H}$}]{
\begin{minipage}[t]{0.2\linewidth}
\centering
\includegraphics[width=1\linewidth]{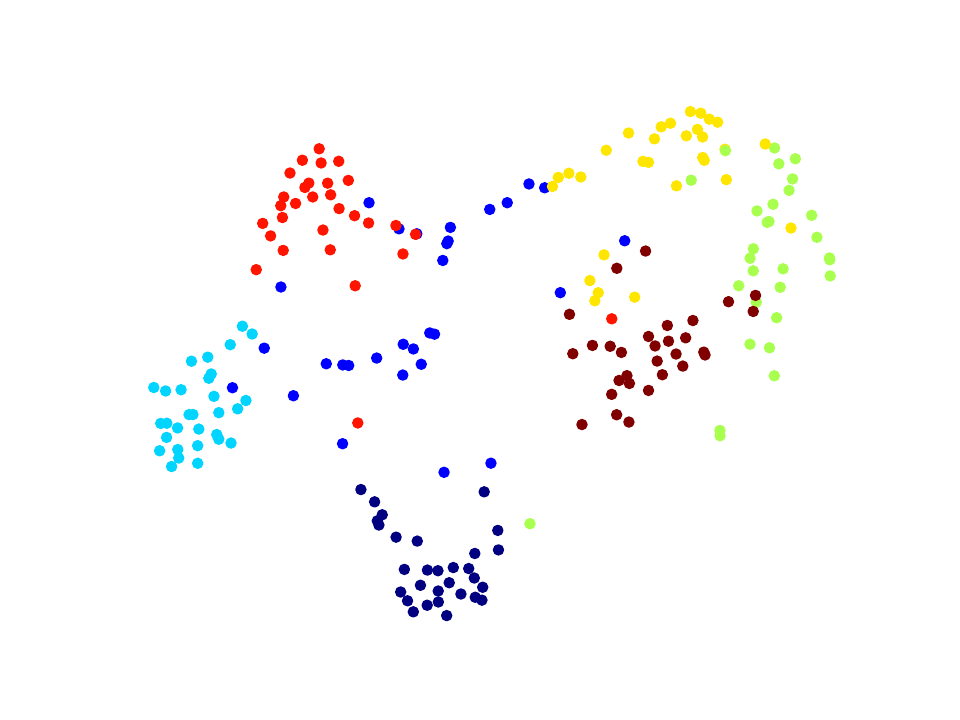}
\end{minipage}%
}%
\subfigure[SLRL-$\tilde{\bm{H}}$]{
\begin{minipage}[t]{0.2\linewidth}
\centering
\includegraphics[width=1\linewidth]{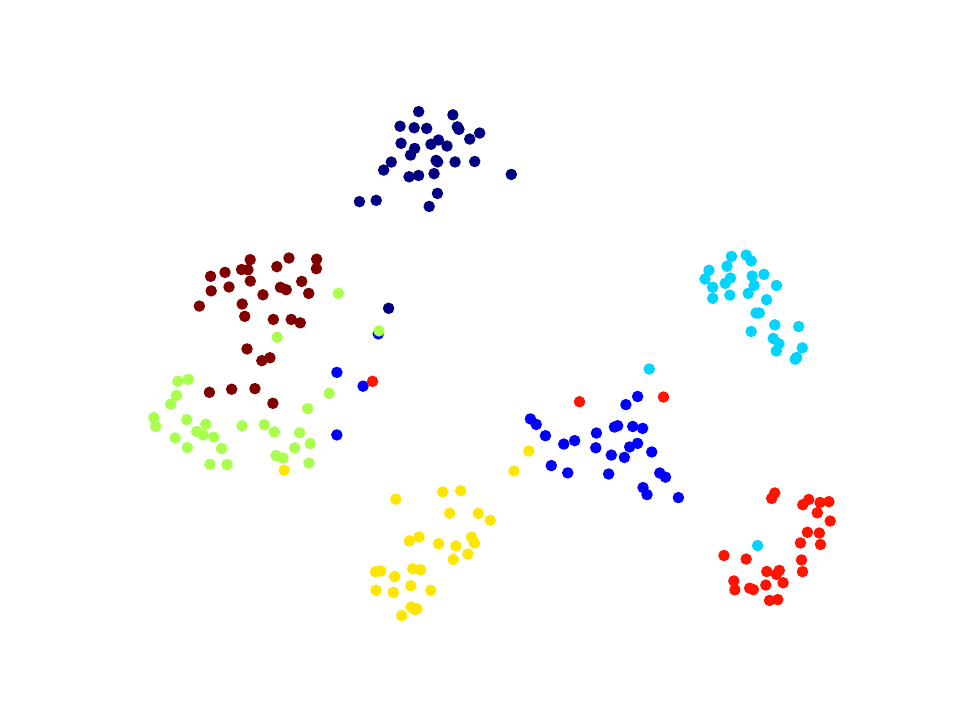}
\end{minipage}%
}%

\centering
\caption{t-SNE visualization of various methods on MSRCV1.}
\label{fig:tsne1}
\end{figure}

To provide a more intuitive visualization, we present the t-SNE visualization results of six methods applied to the MSRCV1 dataset in Fig.\ref{fig:tsne1}. SLRL-$\bm{H}$ represents the visualization of the common latent representation $\bm{H}$, while SLRL-$\tilde{\bm{H}}$ illustrates the visualization of the structured latent representation $\tilde{\bm{H}}$. Observations from Fig.\ref{fig:tsne1} indicate that the latent representations learned by our proposed method demonstrate superior clustering performance compared to those of other methods. Additionally, compared to SLRL-$\bm{H}$, SLRL-$\tilde{\bm{H}}$ shows samples from the same class more tightly grouped and those from different classes more distinctly separated, suggesting a more effective clustering structure. This underscores the efficacy of incorporating graph representation learning and clustering constraints into our model, enabling it to learn features that are more conducive to clustering.

\subsection{Parameter Sensitivity Experiments}
\begin{figure}[htbp!]
\centering
\subfigure[100Leaves]{
\begin{minipage}[t]{0.3\linewidth}
\centering
\includegraphics[width=1\textwidth]{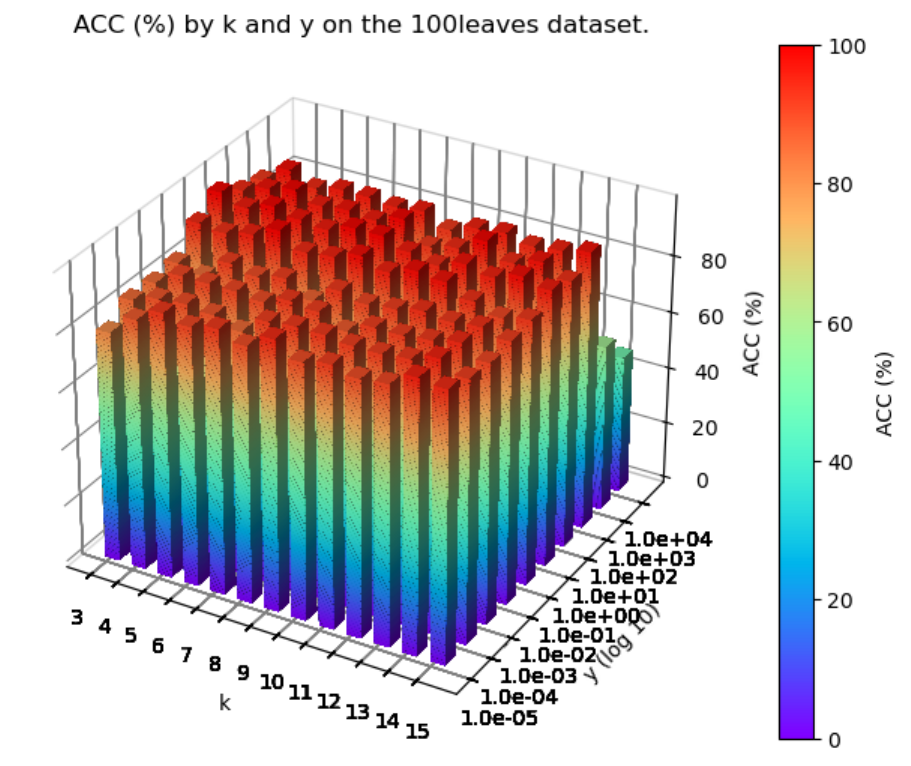}
\end{minipage}}
\subfigure[MSRCV1]{
\begin{minipage}[t]{0.3\linewidth}
\centering
\includegraphics[width=1\textwidth]{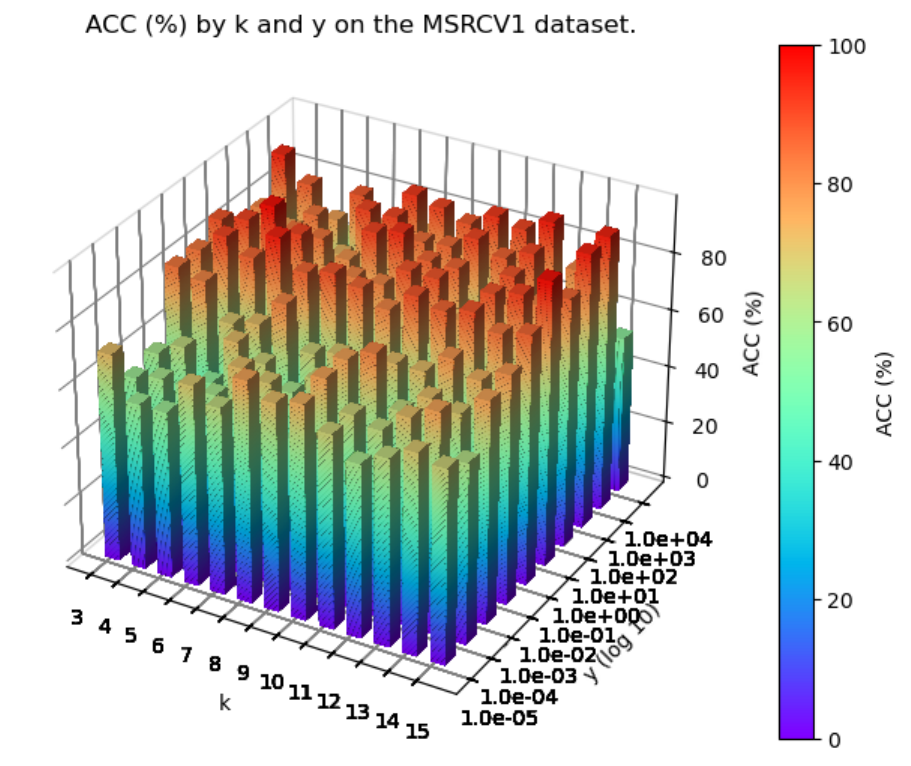}
\end{minipage}}
\subfigure[HW]{
\begin{minipage}[t]{0.3\linewidth}
\centering
\includegraphics[width=1\textwidth]{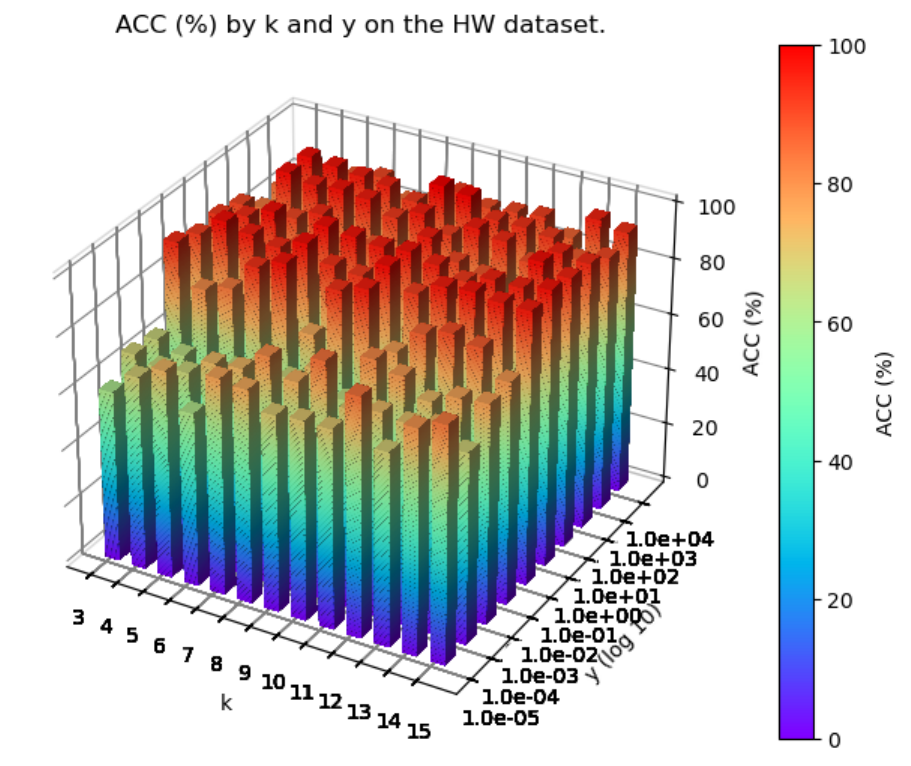}
\end{minipage}}
\caption{Sensitivity experiments for parameters $k$ and $\gamma$.}
\label{fig:parmeter study}
\end{figure}

In our methodology, the number of nearest neighbors $k$ is crucial for constructing the $k$-nearest neighbor graph and significantly influences most graph-based algorithms. Meanwhile, $\gamma$ in the objective function balances the learning of the common representation and clustering constraints. We rigorously evaluated the impacts of $\gamma$ and $k$ on our method's performance, varying $\gamma$ from $10^{-5}$ to $10^{4}$ and adjusting $k$ between 3 and 15. Figure \ref{fig:parmeter study} visualizes these sensitivities across three datasets, showing that the model is robust to $k$ variations within a certain range and performs better when $\gamma > 1$. Overall, our method's performance shows low sensitivity to parameter selection, ensuring reliability and consistency in various scenarios.

\subsection{Convergence Experiments}
\begin{figure}[htbp!]
\centering
\subfigure[100Leaves]{
\begin{minipage}[t]{0.25\linewidth}
\centering
\includegraphics[width=1\textwidth]{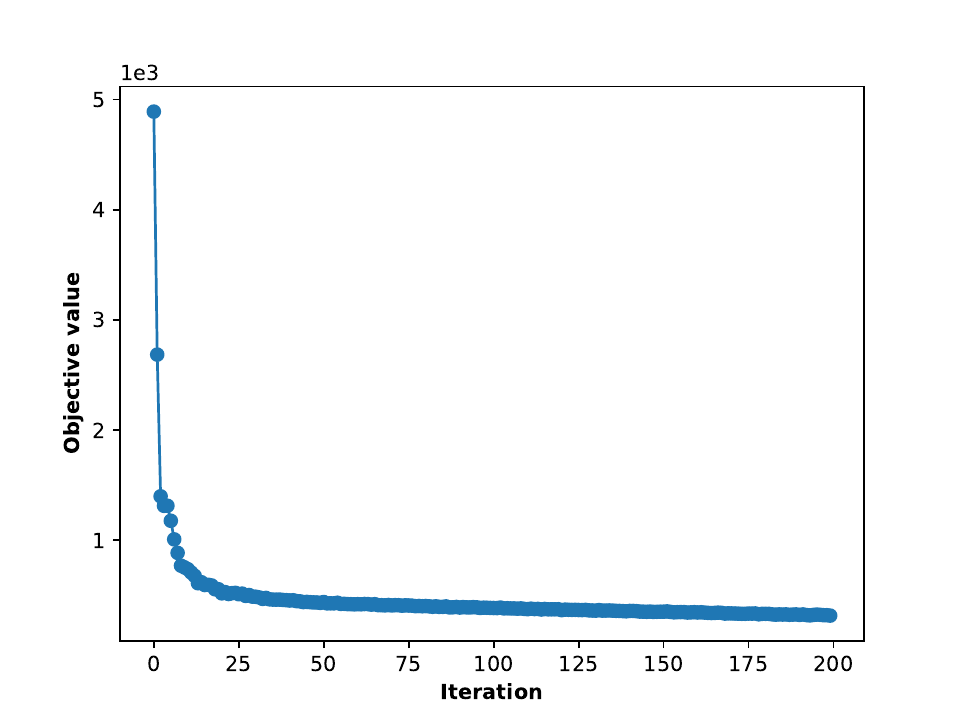}
\end{minipage}}
\subfigure[Scene-15]{
\begin{minipage}[t]{0.25\linewidth}
\centering
\includegraphics[width=1\textwidth]{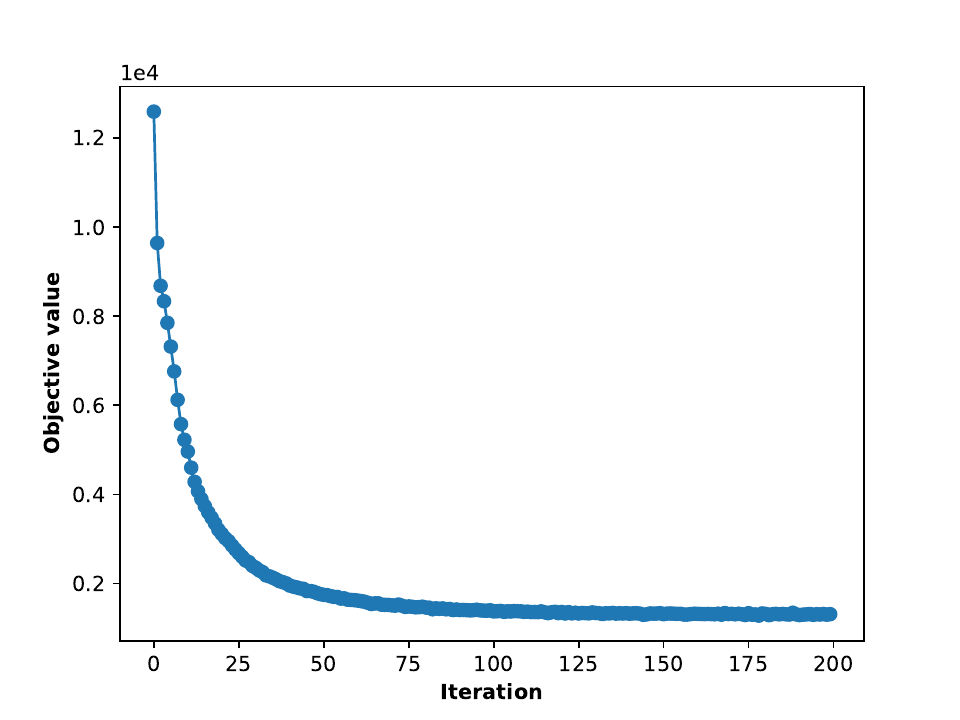}
\end{minipage}}
\subfigure[MSRCV1]{
\begin{minipage}[t]{0.25\linewidth}
\centering
\includegraphics[width=1\textwidth]{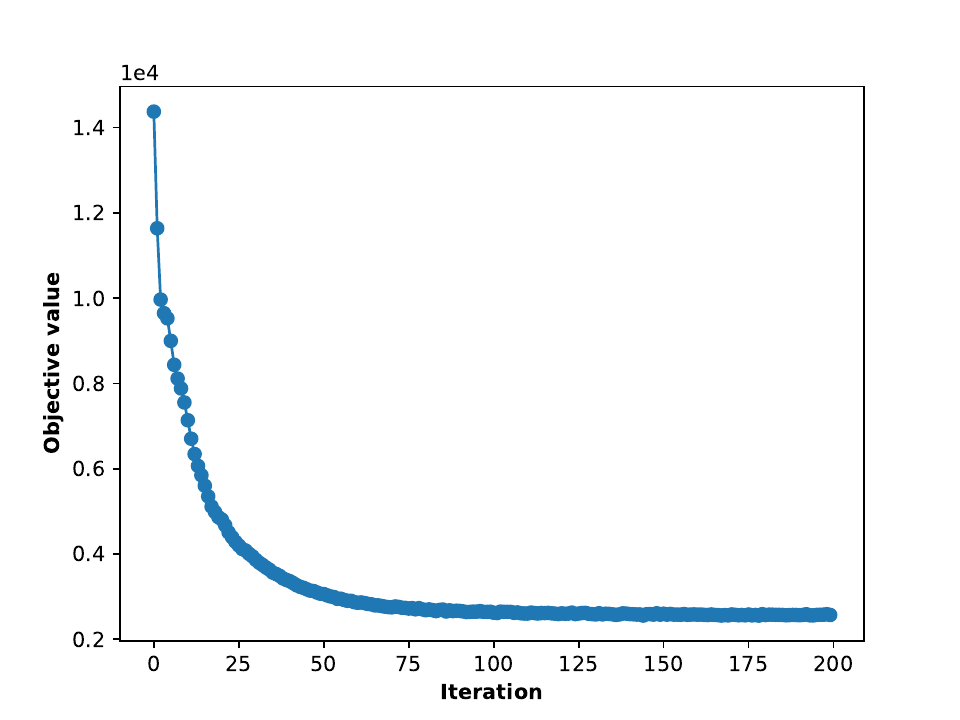}
\end{minipage}}
\subfigure[BBCSport]{
\begin{minipage}[t]{0.25\linewidth}
\centering
\includegraphics[width=1\textwidth]{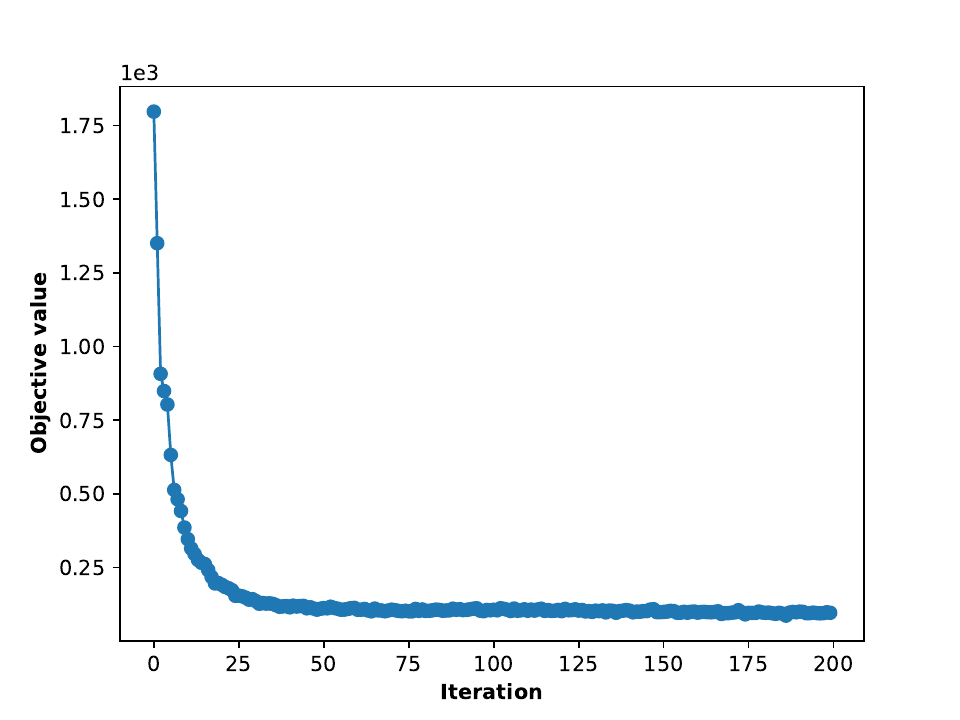}
\end{minipage}}
\subfigure[3Sources]{
\begin{minipage}[t]{0.25\linewidth}
\centering
\includegraphics[width=1\textwidth]{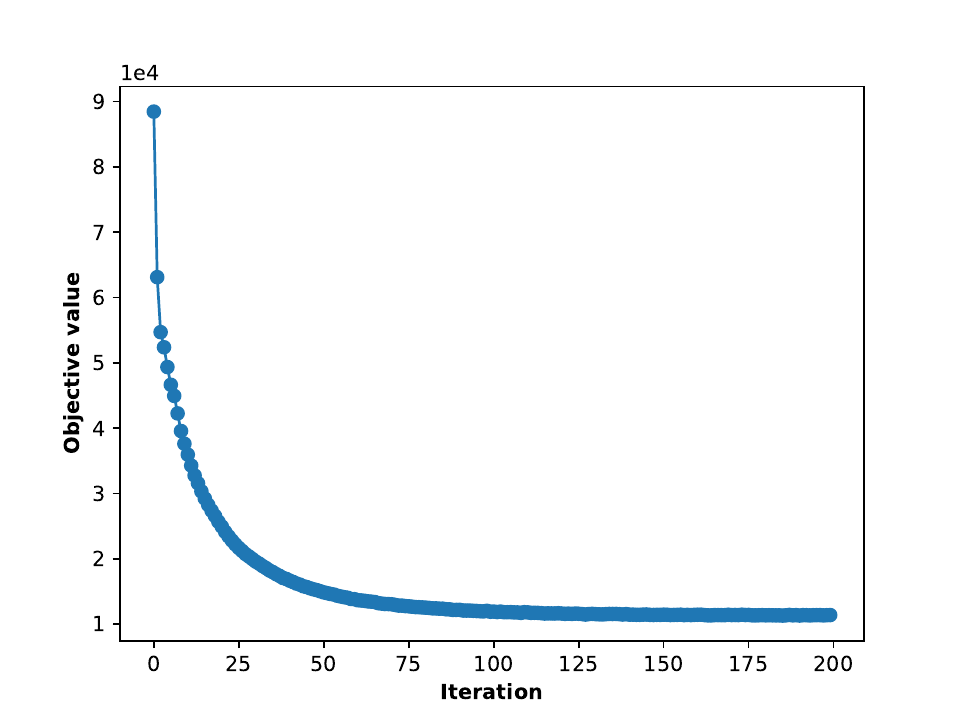}
\end{minipage}}
\subfigure[HW]{
\begin{minipage}[t]{0.25\linewidth}
\centering
\includegraphics[width=1\textwidth]{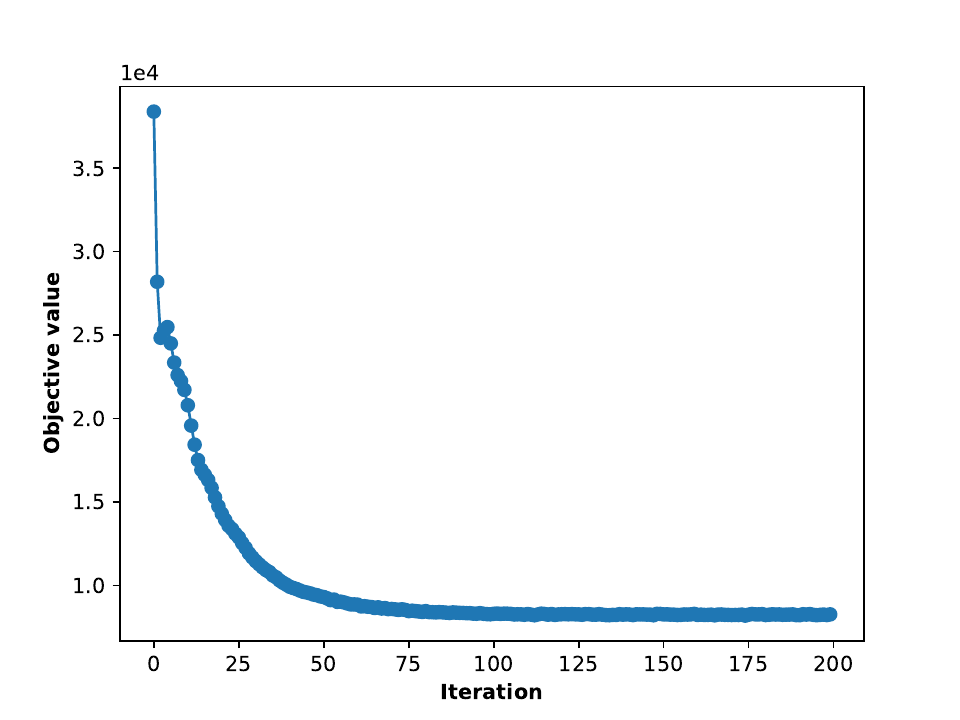}
\end{minipage}}
\caption{Convergent behavior of SLRL.}
\label{fig:curves}
\end{figure}

The convergence curves, depicted in Figure \ref{fig:curves}, illustrate the changes in the values of the total loss function. Initially, there is a rapid decrease in the loss function values within the first few iterations, followed by a slower decline as the number of epochs increases, until convergence is achieved. Typically, for these datasets, our model converges in fewer than 100 epochs, which demonstrates the desirable convergence properties of our proposed method.

\subsection{Ablation Study}
To further validate our proposal, we conducted contrastive learning experiments across various network structures, as documented in Table 2. The following observations can be made:

(a) The common latent representation $\bm{H}$, learned directly from the input features $\bm{X}$ and focusing solely on consistency across different views, did not yield optimal performance.

(b) By incorporating the relationships between different nodes based on $\bm{H}$, we derived a structured latent representation ${\tilde{\bm{H}}}$, which significantly improved performance.

(c) By optimizes both the learning of latent representation and clustering simultaneously, we can obtain a structured latent representation that is more suitable for clustering tasks.

\begin{table}[htbp!]\small
\caption{Results across various network structures.}\label{ablation}
\centering
\begin{tabular}{|l|l|cccccc|}
\hline & & 100Leaves & Scene-15 & MSRCV1 & BBCSport & 3Sources & HW\\  
\hline (a) & $\mathbf{X}-\mathbf{H}_{\checkmark}$ & 80.30 & 35.91 & 71.43 & 89.16 & 72.50 &76.19\\    
\hline (b) & $\mathbf{X}-\mathbf{H}_{\checkmark}-\mathbf{\tilde{\bm{H}}}_{\checkmark}$ & 87.44 & 45.23 & 85.71 & 86.49 & 77.52 &95.23 \\    
\hline (c) & $\mathbf{X}-\mathbf{H}_{\checkmark}-\mathbf{\tilde{\bm{H}}}_{\checkmark}-\mathbf{P}_{\checkmark}$ & \textbf{94.15} & \textbf{47.52} & \textbf{92.70} & \textbf{93.02} & \textbf{81.62} & \textbf{96.40} \\    
\hline    
\end{tabular}    
\end{table}

\section{Conclusion}

This paper introduces a multi-view clustering method based on SLRL, which considers both the complementary relationships between views and the structural relationships among samples, thereby learning a structured latent representation more suitable for clustering tasks. Specifically, to mine the structural information among samples, SLRL constructs a nearest neighbor graph on the common latent representation of all views and employs a graph attention network for graph representation learning, enhancing the structure of the latent representation. Furthermore, a distribution-based clustering loss is employed to further constrain both the common representation learning and graph representation learning, resulting in a latent representation with an improved cluster structure, making it more suitable for clustering tasks. Experimental results on six multi-view datasets demonstrate that the clustering method based on SLRL proposed in this paper achieves commendable clustering performance.
%
%
%
\bibliographystyle{splncs04}
\bibliography{sample-base}
\end{document}